\documentclass[12pt]{article}                     

\usepackage{fullpage}
\usepackage{doi}

\usepackage{graphicx}
\usepackage[ruled]{algorithm2e}
\usepackage{multirow}
\usepackage{mathptmx}      
\usepackage{graphicx}
\usepackage{amsfonts}
\usepackage{algorithm2e}
\usepackage{caption,subcaption}
\usepackage{gnuplottex}
\usepackage{amsthm}
\usepackage{amsmath}
\usepackage{xspace}

\newcommand{\cev}[1]{\reflectbox{\ensuremath{\vec{\reflectbox{\ensuremath{#1}}}}}}
\newcommand{\MMFSA}{\textsc{MaxMinFair\_SA}\xspace}
\newcommand{\LEN}{{\ensuremath n}\xspace}

\makeatletter
\newdimen\@@@tmpa
\newdimen\@@@tmpb
\newdimen\@@@tmpc
\newdimen\@@@tmpd
\def\clap#1#2#3{
       \setbox0=\hbox{#1}\setbox1=\hbox{#2}%
       \@@@tmpa\wd0\@@@tmpb\wd1\advance\@@@tmpa-\@@@tmpb%
       \ifdim\@@@tmpa>0pt\@@@tmpb\wd0%
       \else\@@@tmpb\wd1\fi\@@@tmpc\ht0\@@@tmpd\ht1%
       \advance\@@@tmpc\dp0\advance\@@@tmpd\dp1%
       \advance\@@@tmpc-\@@@tmpd\divide\@@@tmpc2%
       \ifdim\@@@tmpc>0pt\@@@tmpd\ht0\advance\@@@tmpd\dp0\@@@tmpc-\dp0%
       \else\@@@tmpd\ht1\advance\@@@tmpd\dp1\advance\@@@tmpc-\dp0\fi%
       \ifx#3\empty\else\advance\@@@tmpc#3\fi%
       \leavevmode\raise\@@@tmpc\hbox to \@@@tmpb{\rlap{\hbox to \@@@tmpb{\hss%
            \vbox to \@@@tmpd{\vss\box0\vss}\hss}}%
            \hss\vbox to \@@@tmpd{\vss\box1\vss}\hss}%
       }

\makeatother

\newcommand{\T}{{\ensuremath \vartheta}\xspace}
\newcommand{\TT}{{\ensuremath \tau}\xspace}

\newtheorem{definition}{Definition}

\begin{document}

\title{Fairness in Academic Course Timetabling\thanks{Research funded in parts by the School of
Engineering of the University of Erlangen-Nuremberg.
A preliminary version appeared in: Proc. 9th Int. Conf. on the Practice and Theory of
Automated Timetabling (PATAT), 2012, pp. 114--130.}
}

\author{Moritz M\"uhlenthaler\qquad Rolf Wanka \\[2mm]
              Department of Computer Science \\
              University of Erlangen-Nuremberg,
              Germany \\
              \texttt{\{moritz.muehlenthaler,rwanka\}@cs.fau.de}
}

\date{ }

\maketitle

\begin{abstract}

We consider the problem of creating \emph{fair} course timetables in the setting of a
university. Our motivation is to improve the overall satisfaction of individuals
concerned (students, teachers, etc.) by
providing a fair timetable to them. The central idea is that undesirable
arrangements in the course timetable, i.\,e., violations of soft constraints,
should be distributed in a fair way among the individuals. We propose two
formulations for the fair course timetabling problem that are based on max-min
fairness and Jain's fairness index, respectively. Furthermore, we present and
experimentally evaluate an optimization algorithm based on simulated annealing for solving
max-min fair course timetabling problems.
The new contribution is
concerned with measuring the energy difference between two timetables, i.\,e., how
much worse a timetable is compared to another timetable with respect to max-min
fairness. We introduce three different energy difference measures and evaluate
their impact on the overall algorithm performance. The second proposed problem
formulation focuses on the tradeoff between fairness and the total amount of
soft constraint violations. Our experimental evaluation shows that the known best solutions
to the ITC2007 curriculum-based course timetabling instances are quite fair with
respect to Jain's fairness index. However, the experiments also show that
the fairness can be improved further
for only a rather small increase in the total amount of soft constraint
violations.

\end{abstract}


\bigskip

\section{Introduction}
\label{sec:intro}

The university course timetabling problem (UCTP) captures the task of assigning
courses to a limited number of resources (rooms and timeslots) in the setting
of a unversity. In this work, we consider the problem of creating fair course
timetables in the context of a particular variant of the UCTP, the
curriculum-based course timetabling (CB-CTT) problem proposed
in~\cite{ITC2007:CB-CTT}. The CB-CTT formulation features various hard and soft
constraints which model typical real-world requirements. The hard constraints
are basic requirements, for example, no two lectures may be held in the same
room at the same time. The feasible solutions to a CB-CTT instance are
the timetables satisfying all hard constraints. Soft constraints characterize
properties of a course schedule which are undesirable for the stakeholders. The
quality of a feasible timetable is determined by the extent to which the soft
constraints are violated. A soft constraint violation results in a penalty, and
the task is to find a feasible timetable such that the total penalty is
minimal. Situations may arise however, in which a large part of penalty hits
only a small group of the stakeholders, who would thus receive a poor timetable
in comparison to others. In other words, a timetable may be unfair due to an
unequal distribution of penalty. In this work, we apply fairness criteria to
the CB-CTT problem. Our goal is to achieve a balance of interests between the
stakeholders by imposing fairness conditions on the distribution of penalty and
thus, to improve the overall stakeholder satisfaction.

In general, fairness is of interest whenever scarce resources
are allocated to stakeholders with demands.
In economics for example, the distribution of wealth and income and how to
measure inequality of resource distributions is of major concern, see for example \cite{WelfareEconomics}
and
\cite{Sen:97}. In computer science, fairness is a central theme in design and
analysis of communication protocols (see for instance
\cite{Bartal:02,Bertsekas:92,BFT:11,Jain:84,Kelly:98,Kleinberg:01,OW:04,SB:08}).
In operations research, fairness criteria have been applied for example to the
aircraft landing problem by \cite{SK:08}. In the literature, there is a wealth
of different definitions of how to determine the fairness of a given resource
distribution. For instance, we may consider the total amount of allocated
resources, the outcome for the worst-off stakeholder, the deviation from the
mean allocation, and so forth.  We propose two fair variants of the CB-CTT
problem, which differ with respect to the underlying notion of fairness. Our
first problem forumation, MMF-CB-CTT, is based on lexicographic max-min
fairness, to which we will refer to as max-min fairness for brevity. This
fairness notion often appears in the context of network bandwidth allocation
(see for example \cite{Bertsekas:92,SB:08}). Max-min fairness is a purely
qualitative measure of fairness, i.\,e., given two allocations, max-min
fairness tells us which of the two is better, but not by how much. Our second
problem formulation, JFI-CB-CTT, is based on Jain's fairness index proposed by
\cite{Jain:84}.  This fairness measure is used in the famous AIMD algorithm by
\cite{CJ:89} used in TCP Congestion Avoidance. In contrast to max-min fairness,
it conveniently represents the inequality of a resource allocation as a
number between zero and one.

In order to solve the MMF-CB-CTT problem we propose \MMFSA, an optimization
algorithm based on simulated annealing (SA). Due to the mild requirements of SA
on the problem structure, the proposed algorithm can easily be tailored to
other max-min fair optimisation problems. A delicate part of the algorithm is
the energy
difference function, which quantifies how much worse one solution is compared
to another solution -- a piece of information we do not get directly from
max-min fairness. We propose three different energy difference functions and
evaluate their impact on the performance of \MMFSA on the 21 standard instances
from track three of the ITC2007 competition (\cite{ITC2007:web}). Our
experiments indicate, that the known best solutions with respect to the CB-CTT
model are quite fair in the max-min sense, but further improvements are
possible for 15 out of 21 instances and often a considerable improvement of the
worst-off stakeholder is achieved.

The fairness conditions imposed by max-min fairness are rather strict in the
sense that there is no tradeoff between fairness and the total amount of
penalty. In practice however, it may be desirable to pick a timetable from a
number of solutions with varying tradeoffs between fairness and total penalty.
Our proposed JFI-CB-CTT problem is a bi-criteria optimisation problem, which
offers this option. We investigate the tradeoffs between fairness and total
penalty for the six standard instances from~\cite{ITC2007:web} whose known best
solutions have the highest total penalty compared to the other instances. Our
motivation for this choice of instances is simply that if the total penalty of
a timetable is very small, then there is not much gain for anyone in
distributing the penalty in a fair way. Our conclusion regarding this approach
is that, although the known best solutions for the six instances are already
quite fair, we can improve the fairness further at the cost of only a small
increase in total penalty.  For a theoretical treatment of the price of
fairness on so-called convex utility sets with respect to proportional fairness
and max-min fairness, see the recent work by \cite{BFT:11}. 

The remainder of this work is organized as follows. In
Section~\ref{sec:preliminaries}, we will provide a brief review of the
curriculum-based course timetabling (CB-CTT) problem model as well as max-min
fairness and Jain's fairness index. In Section~\ref{sec:fairtimetabling} we
will propose MMF-CB-CTT and JFI-CB-CTT, two fair variants of the CB-CTT problem
formulation, and in Section~\ref{sec:sa}, we will introduce the SA-based
optimisation algorithm \MMFSA for solving max-min fair allocation problems.
Section~\ref{sec:evaluation} is dedicated to our experimental evaluation of the
fairness of the known best solutions to 21 standard instances from the website
by \cite{ITC2007:web} with respect to max-min fairness and Jain's fairness
index, and the performance of the \MMFSA algorithm.


\section {Preliminaries}
\label{sec:preliminaries}

In this section, we provide a brief review of the curriculum-based course
timetabling problem formulation as well as relevant definitions concerning
max-min fairness and Jain's fairness index.

\subsection {Curriculum-based Course Timetabling Problems}
\label{sec:cb-ctt}

Curriculum-based Course Timetabling (CB-CTT) is a particular variant of the
UCTP. It has been proposed in the course of the second international
timetabling competition in 2007 (see \cite{ITC2007}), and has since then
emerged as one of the de-facto standard problem formulations in the timetabling
community. The central entities in the CB-CTT formulation are the
\emph{curricula}, which are sets of lectures that must not be taught
simultaneously. Both problem formulations proposed in the next section are
based on the CB-CTT model.

CB-CTT problems are NP-hard and a lot of effort has been devoted to the
development of exact and heuristic methods which provide high quality solutions
within reasonable time. A wide range of techniques has been employed for
solving CB-CTT instances including but not limited to approaches based on
Max-SAT (\cite{Asin:10}), mathematical programming (\cite{Lach:10,Marecek:11}),
local search (\cite{Gaspero:06,ATS:10}), evolutionary computation
(\cite{Abdullah:07}) as well as hybrid approaches (\cite{BGS:12,Muller:09}). There has
been a lot of progress in terms of the achieved solution quality in the recent
years.  Interestingly, there seems to be no approach which is superior to
the other approaches on most ITC2007 instances (see the website~\cite{ITC2007:web} for current results). 

A CB-CTT instance consists of a set of rooms, a set of courses, a set of
curricula, a set of teachers and a set of days. Each day is divided into a
fixed number of timeslots, a pair composed of a day and a timeslot is called a
\emph{period}. A period in conjunction with a room is called a \emph{resource}.
Each course consists of a number of lectures, i.\,e. a number
events to be scheduled, is taught by some teacher and has a number of students
attending it. Each \emph{curriculum} is a set of courses. For each room, we are
provided with the maximum number of students it can accommodate
and for each course we are given a list of periods in which it cannot be
taught. A solution to a CB-CTT instance is a \emph{timetable}, i.\,e. an
assignment of courses to resources. The quality of a timetable is determined
according to four hard and four soft constraints (see \cite{ITC2007:CB-CTT}).

The hard constraints are the following:
\begin {enumerate}
\item[H1\label{itm:H1}]
All lectures need to be scheduled and no two lectures of the same course may be assigned to the same period.
\item[H2\label{itm:H2}]
No two lectures may be assigned to the same resource.
\item[H3\label{itm:H3}]
Two courses in the same curriculum or taught by the same teacher must be assigned to different periods.
\item[H4\label{itm:H4}]
A lecture can only be scheduled in a period that is not marked unavailable for
the corresponding course.
\end {enumerate}
A timetable that satisfies all hard constraints is called \emph{feasible}.

The CB-CTT formulation features the following soft constraints:
\begin {enumerate}
\item[S1\label{itm:S1}] \emph{RoomCapacity}:
Each lecture should be assigned to a
room of sufficient size. 
\item[S2\label{itm:S2}] \emph{MinWorkingDays}:
The lectures of each course should 
be distributed over a certain minimum number of days.
\item[S3\label{itm:S3}] \emph{IsolatedLectures}:
For each curriculum, all lectures
associated to the curriculum should be scheduled in adjacent timeslots.
\item[S4\label{itm:S4}] \emph{RoomStability}:
The lectures of each course should be assigned to the same room.
\end {enumerate}

Each violation of one of the soft constraints results in a ``penalty'' for the
timetable. The CT-CTT objective function aggregates individual penalties by
taking their weighted sum. Detailed descriptions of how hard and soft
constraints are evaluated and how much penalty is applied for a particular soft
constraint violation can be found in the report by \cite{ITC2007:CB-CTT}. 
Given a CB-CTT instance $I$, the task is to find a feasible 
timetable such the aggregated penalty is minimal.

\subsection {Fairness in Resource Allocation}
\label{sec:fairresources}

Fairness issues typically arise when scarce resources are allocated to a number
of stakeholders with demands. Fair resource allocation has received much
attention in economic theory (see for example \cite{WelfareEconomics}), but also occurs in a wide
range of applications in computer science including bandwidth allocation in
networks (\cite{Bertsekas:92}) and task scheduling (\cite{RDK:05}). In many
optimization problems related to resource allocation, the goal is to maximize
the total amount of resources allocated to the stakeholders. Fairness in this
context means that the distribution of resources over the stakeholders is
important and that certain resource distributions are preferred over others.

Consider a resource allocation problem with $n$ stakeholders resources. Each
resource allocation (admissible solution) induces an allocation vector $X =
(x_1, \ldots, x_n)$, where each item $x_i, 1 \leq i \leq n$, corresponds to the
amount of resources allocated to stakeholder $i$. Typically, a preference for
certain resource distributions is implicitly or explicitly contained in the
objective function. For example, the task can be to find allocations
maximizing the sum of the individual allocations, the mean allocation, the root
mean square (RMS), the smallest allocation, and so
forth (see \cite{Ogryczak:10,SK:08}). When the fairness of an allocation is important
we may be interested in improving the outcome for the worst-off stakeholders or
generally try to allocate resources equally among the stakeholders.
Max-min fairness is a notion of fairness that favours better outcomes for the
worst-off stakeholders. It has received attention in the area of network
engineering, in particular in the context of flow
control~\cite{Bartal:02,Kleinberg:01,SB:08,Zhang:11}. Various inequality
measures have been proposed such as the Gini index proposed by 
\cite{Gini:21} and Jain's fairness index proposed by \cite{Jain:84}. Generally,
a highly unequal distribution of resources is considered unfair. Our evaluation
of fairness in academic course timetabling focuses on max-min fairness and
Jain's fairness index.

Our evaluation of fairness in academic course timetabling focuses on the two
fairness criteria max-min fairness and Jain's fairness index.

\paragraph{Max-min Fairness.}
Max-min
fairness can be stated as iterated application of Rawls's Second Principle
of Justice by \cite{Rawls:2005}: 
\begin{quote}
``Social and economic inequalities are to be arranged so that they are to be of
greatest benefit to the least-advantaged members of society.'' (the Difference
Principle)
\end{quote}
Once the status of the least-advantaged members has been determined according
to the difference principle, it can be applied again to everyone except the
least-advantaged group in order to maximize the utility (in the economic sense)
for the second least-advantaged members, and so on. The resulting utility
assignment is called max-min fair. A max-min fair utility assignment implies
that each stakeholder can maximize his/her utility as long as it is not at the
expense of another stakeholder who is worse off. Thus, a max-min fair
allocation enforces an efficient resource usage to some extent. A max-min fair
resource allocation is Pareto-optimal. 

In order to define max-min fairness more formally, we introduce some notation.
Let $X$ be an allocation vector. We generally assume that each entrie of $X$
is a nonnegative real number. By $\vec X$ we denote the vector containing the entries of $X$
arranged in nondecreasing order. Similarly, let $\cev X$ be a
vector containing the entries of $X$ in nonincreasing order. For allocation
vectors $X$ and $Y$ we write $X \preceq_{\it mm} Y$ if $X$ is at least as good
as $Y$ in the max-min sense. For maximization problems such as bandwidth
allocation this is the case iff $\vec Y \preceq_{\it lex} \vec X$, where
$\preceq_{\it lex}$ is the usual lexicographic comparison. For minimization
problems such as the fair course timetabling problems proposed in
Section~\ref{sec:fairtimetabling}, $X \preceq_{\it mm} Y$ iff $\cev X
\preceq_{\it lex} \cev Y$. Let $s$ be a solution to an instance $I$ of an
optimization problem and $X$ be the allocation vector induced by $s$. Then $s$
is called max-min fair, if for any other solution $s'$ to $I$ we have $X
\preceq_{\it mm} Y$, where $Y$ is the allocation vector induced by $s'$.
Since the allocations are sorted, max-min fairness does not discriminate
between stakeholders, but only between the amounts of resources assigned to
them. 

A weaker version of max-min fairness results if Rawl's Second Principle of
Justice is not applied iteratively, but just once. This means that we are
concerned with chosing the best possible outcome for the worst-off
stakeholder. In the literature, related optimization problems are referred
to as bottleneck optimization problems (\cite{EF:70,PZ:11}). Note, that in
contrast to max-min fairness, an optimal solution to a bottleneck optimisation
problem is not necessarily Pareto-optimal. In the context of practical academic
timetabling, the use of bottleneck optimization is hard to justify: each
stakeholder is guaranteed to be at least as well off as the worst-off
stakeholder, but no further improvement is considered. 

\paragraph{Jain's Fairness Index.}
Jain's fairness index is an inequality measure proposed by Jain in~\cite{Jain:84}. It
is the crucial fairness measure that is used in the famous AIMD algorithm
by~\cite{CJ:89} used in TCP Congestion Avoidance. The fairness index~$J(X)$ of
an allocation vector $X$ is defined as follows:
\begin{equation}
	\label{eq:jain}
	J(X) = \frac{\displaystyle\Big(\sum_{1\leq i \leq n} x_i\Big)^2}
                    {\displaystyle n\cdot \sum_{1\leq i \leq n} x_i^2}
\enspace.
\end{equation}
It has several useful properties like population size independence, scale and
metric independence, it is bounded between $0$ and $1$, and it has an intuitive
interpretation. In particular $J(X)=1$ means that $X$ is a completely fair
allocation, i.\,e., the allocation is fair for every stakeholder, and if
$J(X)=1/n$ then all resources are occupied by a single stakeholder. Furthermore, 
if $J(X)=x{\,\%}$ then the allocation $X$ is fair for $x$ percent of the
stakeholders.


\section {Fairness in Academic Course Timetabling}
\label {sec:fairtimetabling}

Course timetabling problems fit quite well in the framework of fair resource
allocation problems described in the previous section: A timetable is an
allocation of resources (rooms, timeslots) to lectures. In this section, we will
define two fair versions of the CB-CTT problem formulation proposed
by \cite{ITC2007}. The first one, MMF-CB-CTT, is based on max-min fairness.
Since max-min fairness enforces fairness as well as efficiency (maximum
utility) at least to some extent, it is not a suitable concept for exploring
the tradeoff between fairness and efficiency. Thus, we propose a second fair
variant of CB-CTT called JFI-CB-CTT that is based on Jain's fairness index.

In order to use the fairness measures mentioned in the previous section, we
need determine an allocation vector from a timetable. The central entities in
the CB-CTT problem formulation are the curricula.  Therefore, in this work, we
are interested in a fair distribution of penalty over the curricula. Depending
on the application, a different set of stakeholders can be picked, but
conceptually this does not change much. Let $I$ be a CB-CTT instance with
curricula $c_1, c_2, \ldots, c_k$ and let $f_c$ be the CB-CTT objective
function proposed by \cite{ITC2007:CB-CTT}, which evaluates (S1)-(S4)
restricted to curriculum $c$. This means $f_c$ determines soft constraint
violations only for the courses in curriculum $c$. For a timetable \TT the
corresponding allocation vector is given by the allocation function 
\begin{equation}
\label{eq:allocation}
A(\TT) = (f_{c_1}(\TT), f_{c_2}(\TT), \ldots, f_{c_k}(\TT))
\enspace.
\end{equation}

\begin{definition}[MMF-CB-CTT] Given a CB-CTT instance $I$, the task is to find
a feasible timetable $\TT$ such that $A(\TT)$ is max-min fair. 
\end{definition}

If a feasible timetable has max-min fair allocation vector, then any curriculum
$c$ could receive less penalty only at the expense of other curricula which
receive more penalty than $c$. This is due to the Pareto-optimality of a
max-min fair allocation. 

In order to explore the tradeoff between efficient and fair resource allocation
in the context of the CB-CTT model, we propose another fair variant called
JFI-CB-CTT that is based on Jain's fairness index proposed by \cite{Jain:84}. In
order to get meaningful results from the fairness index however, we need a different
allocation function. Consider an allocation $X$ that allocates all penalty 
to a single curriculum while the remaining $k-1$ curricula receive no penalty.
Then $J(X) = 1/k$, which means that only one curriculum is happy with the
allocation (see \cite{Jain:84}). In our situation however, the opposite is the case:
$k-1$ curricula are happy since they receive no penalty at all. The following
allocation function shifts the penalty values such that the corresponding
fairness index in the situation described above becomes $(k-1)/k$, which is in
much better agreement with our intuition:
\begin{equation}
\label{eq:shiftallocation}
A'(\TT) = (f_{\it max} - f_{c_1}(\TT), f_{\it max} - f_{c_2}(\TT), \ldots, f_{\it max}
- f_{c_k}(\TT))
\enspace,
\end{equation}
with
$$
	f_{\it max} = \max_{1 \leq i \leq k} \left\{ f_{c_i}(\TT) \right\}
	\enspace.
$$

\begin{definition}[JFI-CB-CTT]
Given a CB-CTT instance $I$, the task is to find the set of feasible solutions
which are Pareto-optimal with respect to the two objectives of the objective function 
\begin{equation}
F(\TT) = (f(\TT), 1-J(A'(\TT)))
\enspace,
\end{equation}
where $f$ is the CB-CTT objective function from~\cite{ITC2007:CB-CTT}
and $J$ is defined in Eq.~\eqref{eq:jain}.
\end{definition}

By a similar procedure, other classes of timetabling problem such as post-enrollment
course timetabling, exam timetabling and nurse rostering can be turned into
fair optimization problems. For example, for post-enrollment course
timetabling, the central entities of interest are the individual
students. Therefore, the goal were to achieve a fair distribution of penalty
over all students. Once an appropriate allocation function has been defined,
we immediately get the corresponding fair optimization problems.

Our proposed problem formulations are concerned with balancing the interests
between stakeholders, who are in our case the students. In practice however, there
are often several groups of stakeholders with possibly conflicting interest, for
example students, lecturers and administration.  possibilities for extending
the problem formulations above to include multiple stakeholders. For example, a
multi-objective optimization approach may be considered, where each objective
captures the fairness with respect to a particular stakeholder. When using
inequality measures like Jain's fairness index for different groups of
stakeholders, the inequality values can be aggregated, for instance using a
weighted-sum or ordered weighted averaging~\cite{Yager:88} approach.
Furthermore, max-min fairness or a suitable inequality measure can be applied
to the different objectives to balance the interests of the different groups of
stakeholders.


\section {Simulated Annealing for Max-Min Fair Course Timetabling}
\label{sec:sa}

Simulated Annealing (SA) is a popular local search method proposed by
\cite{Kirkpatrick:83}, which works surprisingly well on many problem domains.
SA has been applied successfully to timetabling problems by \cite{Kostuch:05}
and \cite{Thompson:96}. Some of the currently known best solutions to CB-CTT
instances from the ITC2007 competition were discovered by simulated
annealing-based methods according to the website by \cite{ITC2007:web}. Our SA for max-min fair
optimization problems shown in Algorithm~\ref{alg:sa} below (algorithm \MMFSA)
is conceptually very similar to the original algorithm. The SA algorithm
generates a new candidate solution according to some neighborhood exploration
method, and replaces the current solution with a certain probability depending
on i) the quality difference between the two solutions and ii) the current
temperature. Since max-min fairness only
tells us which of two given solutions is better, but not how much better, the
main challenge in tailoring SA to max-min fair optimization problems is to find
a suitable energy difference function, which quantifies the difference in
quality between two candidate solutions. In the following, we propose three
different energy difference measures for max-min fair optimization and provide
details on the acceptance criterion, the cooling schedule, and the neighborhood
exploration method chosen for the experimental evaluation of \MMFSA in the next
section.

\begin{algorithm}[]
\caption{\MMFSA} 
\label{alg:sa}

\DontPrintSemicolon

\SetKwData{TCUR}{$\mathrm{s_{cur}}$}
\SetKwData{TNEXT}{$\mathrm{s_{next}}$}
\SetKwData{TBEST}{$\mathrm{s_{best}}$}
\SetKwData{TMAX}{$\mathrm{\T_{max}}$}
\SetKwData{TMIN}{$\mathrm{\T_{min}}$}
\SetKwData{Temp}{$\mathrm{\T}$}
\SetKwData{TIMEOUT}{$\mathrm{timeout}$}

\SetKwInOut{Input}{input}
\SetKwInOut{Output}{output}

\Input{\TCUR: feasible timetable, \TMAX: initial temperature, \TMIN: final temperature, \TIMEOUT}
\Output{\TBEST: Best feasible timetable found so far}

\BlankLine

\TBEST $\leftarrow$ \TCUR \;
\Temp $\leftarrow$ \TMAX \;

\While{timeout not hit}{
	\TNEXT $\leftarrow$ \FuncSty{neighbor(\TCUR)} \;
	\lIf {${P_{\it accept}} \geq$ \FuncSty{random()}} {
		\TCUR $\leftarrow$ \TNEXT \;
	}
	\lIf {$A(\TCUR) \preceq_{\it mm} A(\TBEST)$} {
		\TBEST $\leftarrow$ \TCUR \;
	}
	\Temp $\leftarrow$ \FuncSty{next\_temperature(\Temp)} \;
}

\Return{\TBEST}
\end{algorithm}

\paragraph{Acceptance Criterion.} 
Similar to the original SA algorithm proposed by \cite{Kirkpatrick:83},
algorithm~\MMFSA accepts an improved or equally good solution
$\mathrm{s_{next}}$ with probability 1. If $\mathrm{s_{next}}$ is worse than
$\mathrm{s_{cur}}$ then  the acceptance probability depends on the current
temperature level $\T$ and the energy difference $\Delta E$. The energy
difference measures the difference in quality of the allocation induced by
$\mathrm{s_{next}}$ compared to the allocation induced by the current solution
$\mathrm{s_{cur}}$. The acceptance probability $P_{\it accept}$ is defined as:
$$
	P_{\it accept} = 
	\begin{cases}
		1 & \text{if } \mathrm{s_{next}} \preceq_{\it mm } \mathrm{s_{cur}} \\
		\exp\big(-\dfrac{\Delta E(X, Y)}{\T}\big) & \text{otherwise},
	\end{cases}
$$
where $X = A(\mathrm{s_{cur}})$ and $Y = A(\mathrm{s_{next}})$.
In order to fit max-min fairness into the SA algorithm, we propose three energy
difference measures: $\Delta E_{\it lex}$, $\Delta E_{\it cw}$, and $\Delta
E_{\it ps}$. $\Delta E_{\it lex}$ derives the energy difference from a
lexicographic comparison, $\Delta E_{\it cw}$ from the component-wise ratios of
the sorted allocation vectors and $\Delta E_{\it ps}$ from the ratios of the
partial sums of the sorted allocation vectors. Our
experiments presented in the next section indicate that the choice of the
energy difference measure has a clear impact on the performane of \MMFSA and is
thus a critical design choice. 

For an allocation vector $X$, let $\cev X_i$ denote the $i$th entry after
sorting the entries of $X$ in nonincreasing order. The energy difference
$\Delta E_{\it lex}$ of two allocation vectors $X$ and $Y$ of length \LEN is
defined as follows:
\begin{equation}
\label{eq:elex}
\Delta E_{\it lex}(X, Y) = 
 1 - \frac{1}{\LEN} \cdot \left( \min_{1 \leq i \leq \LEN} \left \{ i \mid \cev Y_i > \cev X_i \right \} - 1\right)
\enspace.
\end{equation}
$\Delta E_{\it lex}$ determines the energy difference between $X$ and $Y$ from
the smallest index that determines $X \preceq_{\it mm} Y$. Thus, sorted
allocation vectors which differ at the most significant indices have a higher
energy difference than those differing at later indices. In particular, $\Delta
E_{\it lex}$ has evaluates to $1$ if the minimum is $1$, and it evaluates to
$1/n$ if the minimum is \LEN.

The energy difference measure $\Delta E_{\it lex}$ considers the earliest index
at which two sorted allocation vectors differ but not how much the
entries differ. We additionally propose the two energy difference
measures $\Delta E_{\it cs}$ and $\Delta E_{\it ps}$ which take this
information into account. These energy difference measures are derived from the
definitions of approximation ratios for max-min fair allocation problems given
by \cite{Kleinberg:01}. An approximation ratio is a measure for how much worse
the quality of a solution is relative to a possibly unknown optimal solution.
In our case, we are interested in how much worse one given allocation is
relative to another given allocation. We need to introduce some
modifications of the definitions by \cite{Kleinberg:01} since we are dealing
with a minimization problem.

Note that due to~\eqref{eq:allocation}, an allocation vector does not contain
any positive entries. Let $\mu_{X,Y}$ be the smallest value of the two
allocation vectors $X$ and $Y$ offset by a parameter $\delta >
0$, i.\,e., 
\begin{equation}
\label{eq:mu}
\mu_{X,Y} = \max \{ \cev X_1, \cev Y_1 \} + \delta 
\enspace.
\end{equation}
The offset $\delta$ is introduced in order to avoid divisions by zero when
taking ratios of penalty values.

The component-wise energy difference $\Delta E_{\it cw}$ of allocation vectors
$X$ and $Y$ is defined as follows:
\begin{equation}
\label{eq:ecw}
\Delta E_{\it cw}(X, Y) = \displaystyle\max_{1 \leq i \leq \LEN} \left \{ \frac{\mu_{X, Y} - \cev Y_i}{\mu_{X, Y} - \cev X_i} \right \} - 1
\end{equation}
Since all entries are subtracted from $\mu_{X, Y}$, the ratios of the most
significant entries with respect to $\preceq_{\it mm}$ tend to dominate the
value of $\Delta E_{\it cw}$. Consider for example the situation that $Y$
is much less fair than $X$, say, $\max \{ \vec X_1, \vec Y_1 \}$ occurs more often
in $X$ than in $Y$. Then for a small offset $\delta$ the energy difference $\Delta
E_{\it cw}(X, Y)$ becomes large. On the other hand, if $X$ is nearly as fair as
$Y$ then the ratios are all close to one and thus $\Delta E_{\it cw}(X, Y)$ is
close to zero.

The third proposed energy difference measure $\Delta E_{\it ps}$ is based on
the ratios of the partial sums $\sigma_i(X)$ of the sorted allocation vectors. 
$$
\sigma_i(X) = \displaystyle\sum_{1 \leq j \leq i} \cev X_j
\enspace.
$$
The intention of using partial sums of the sorted allocations is to give the
stakeholders receiving the most penalty more influence on the resulting energy
difference compared to $\Delta E_{\it cw}$. The energy difference $\Delta
E_{\it ps}$ is defined as:
\begin{equation}
\label{eq:eps}
\Delta E_{\it ps}(X, Y) = \displaystyle\max_{1 \leq i \leq \LEN} \left \{ \frac{i\cdot\mu_{X,Y} - \sigma_i(\vec Y)}{i\cdot\mu_{X,Y} - \sigma_i(\vec X)} \right \} - 1
\enspace.
\end{equation}


\paragraph {Cooling Schedule.}
In algorithm~\MMFSA, 
the function \FuncSty{next\_temperature} updates the
current temperature level $\T$ according to the cooling schedule. We use
a standard geometric cooling schedule
$$
	\T = \alpha^t \cdot \T_{max}
        \enspace,
$$
where $\alpha$ is the cooling rate and $t$ is the elapsed time. Geometric
cooling schedules decrease the temperature level exponentially over time. It is
a popular class of cooling schedules which is widely used in practice and works
well in many problem domains including timetabling
problems~\cite{Laarhoven:87,Koulamas:94,TD:1998}. Geometric cooling was chosen
due to its simplicity, since the main focus of our evaluation in
Section~\ref{sec:evaluation} is the performance impact of the different energy
difference functions. We have made a slight adjustment to the specification of
the geometric cooling schedule in order to make the behavior more consistent
for different timeouts. Instead of specifying the cooling rate $\alpha$, we
determine $\alpha$ from $\T_{max}$, the desired minimum temperature $\T_{min}$
and the timeout according to:
\begin{equation}
\label{eq:alpha}
\alpha = \left( \frac{\T_{min}}{\T_{max}}  \right)^{\frac{1}{\mathrm{timeout}}} \enspace.
\end{equation}
Thus, at the beginning ($t=0$) the temperature level is $\T_{max}$ and when the
timeout is reached ($t=\mathrm{timeout}$), the temperature level becomes
$\T_{min}$. We chose to set a timeout rather than a maximum number of iterations
since this setting is compliant with the ITC2007 competition conditions, which
are a widely accepted standard for comparing results.

\paragraph{Neighborhood.}
In our max-min fair SA implementation, the function \FuncSty{neighbor} picks at
random a neighbor in the Kempe-neighborhood of $\mathrm{s_{cur}}$. The
Kempe-neighborhood is the set of all timetables which can be reached by
performing a single Kempe-move such that the number of lectures per period do
not exceed the number of available rooms. The Kempe-move is a well-known and
widely used operation for swapping events in a
timetable~\cite{Burke:2010,ATS:10,MBHS:03,TD:1998,HSA}. A prominent feature of
the Kempe-neighborhood is that it contains only moves that preserve the
feasibility of a timetable. Since the algorithm~\MMFSA only uses moves from
this neighborhood the output is guaranteed to be a feasible timetable. In the
future, more advanced neighborhood exploration methods similar to the
approaches in~\cite{Gaspero:06,ATS:10} could be used, which may well lead to an
improved overall performance of~\MMFSA.

\section {Evaluation}
\label{sec:evaluation}

\begin{table}
\caption{Fairness of the known best timetables from~\cite{ITC2007:web} for the
ITC2007 CB-CTT
instances.\label{tab:fairbest}}
$$\small
\begin{tabular}{@{}|r|c|c|c|l|@{}}
\hline
\small Instance & \small Curricula & \small $f(s_{\it best})$ & \small $J(A'(s_{best}))$ & \small$\vec A(s_{best})$ \rule[-6pt]{0pt}{16pt}\\
\hline
\tt{comp01}     & 14 & 5	& 0.8571	& $5^{2},0^{12}$\rule{0pt}{11pt}\\
\tt{comp02}     & 70 & 24	& 0.9515	& $4,2^{10},0^{59}$\\
\tt{comp03}     & 68 & 66	& 0.9114	& $13,10^{3},9,7^{2},6^{4},5^{13},4,2^{6},0^{37}$\\
\tt{comp04}     & 57 & 35	& 0.8964	& $7,6^{3},5^{4},4^{2},2,0^{46}$\\
\tt{comp05}     & 139 &291	& 0.8277	& $41^{2},36^{7},35^{5},32^{5},31^{6},30^{9},28,27^{7},26^{2},25^{14},\ldots,2,0^{3}$\\
\tt{comp06}     & 70 & 27	& 0.9657	& $12,7^{2},5^{4},2^{3},0^{60}$\\
\tt{comp07}     & 77 & 6	& 0.9870	& $6,0^{76}$\\
\tt{comp08}     & 61 & 37	& 0.9020	& $7,6^{3},5^{4},4^{2},2^{2},0^{49}$\\
\tt{comp09}     & 75 & 96	& 0.8047	& $10^{5},9,7^{10},6^{6},5^{10},4,2,0^{41}$\\
\tt{comp10}     & 67 & 4	& 0.9701	& $2^{2},0^{65}$\\
\tt{comp11}     & 13 & 0	& $-$		& $0^{13}$\\
\tt{comp12}     & 150 &300	& 0.9128	& $45,30^{14},28,27^{2},26^{5},25^{19},22^{4},21^{6},20^{8},19,\ldots,2^{2},0^{3}$\\
\tt{comp13}     & 66 & 59	& 0.8830	& $8,7,6^{5},5^{7},4^{2},2^{3},0^{47}$\\
\tt{comp14}     & 60 & 51	& 0.9023	& $8^{4},7,5^{2},2^{6},0^{47}$\\
\tt{comp15}     & 68 & 66	& 0.8495	& $10^{3},9^{3},7,6^{4},5^{13},4,2^{7},0^{36}$\\
\tt{comp16}     & 71 & 18	& 0.9176	& $7^{2},5^{7},4,0^{61}$\\
\tt{comp17}     & 70 & 56	& 0.9248	& $10^{2},6^{3},5^{9},2^{4},0^{52}$\\
\tt{comp18}     & 52 & 62	& 0.9009	& $17,15,14,13,11,10,9^{2},5^{19},2^{2},0^{23}$\\
\tt{comp19}     & 66 & 57	& 0.9612	& $13,7,6^{4},5^{2},4,2^{7},0^{50}$\\
\tt{comp20}     & 78 & 4	& 0.9744	& $2^{2},0^{76}$\\
\tt{comp21}     & 78 & 76	& 0.8838	& $12,11,10^{4},9,7^{4},6^{4},5^{12},4,2^{3},1^{2},0^{45}$\\
\hline
\end{tabular}
$$
\end{table}

In this section, we will first address the question how fair or unfair the
known best timetables for the ITC2007 CB-CTT instances are with respect to
Jain's fairness index and max-min fairness. Table~\ref{tab:fairbest} shows our
measurements for all instances \texttt{comp01}, \texttt{comp02}, \ldots,
\texttt{comp21} from the ITC2007 competition (see \cite{ITC2007:web} for
instance data). Please note that the known best timetables were not created
with fairness in mind, but the objective was to
create timetables with minimal total penalty. In Table~\ref{tab:fairbest},
$s_{\it best}$ refers to the known best solution for each instance. $A$ and
$A'$ refer to the allocation functions given in~\eqref{eq:allocation}
and~\eqref{eq:shiftallocation}, respectively. The data indicates that the
timetables with a low total penalty are also rather fair. This can be explained
by the fact that these timetables do not have a large amount of penalty to
distribute over the curricula. Thus, most curricula receive little or no
penalty and consequently, the distribution is fair for most curricula. We will
show in Section~\ref{sec:tradeoff} however, that for timetables with a
comparatively large total penalty there is still some room for improvement
concerning fairness. 

The rightmost column of Table~\ref{tab:fairbest} contains the sorted allocation
vectors of the best solutions. For a more convenient presentation, all
entries of the sorted allocation vectors are multiplied by -1. The exponents
denote how often a certain number occurs. For example, the sorted allocation
vector $(-5, -5, 0, 0, 0)$ would be represented as $5^{2},0^{3}$. The
sum of the values of an allocation vector is generally much larger than the
total penalty shown in the second column. The reason for this is that the
penalty assigned to a course is counted for each curriculum the course belongs
to. With a few exceptions the general theme seems to be that the penalty is
assigned to only a few curricula while a majority of curricula receives no
penalty. In the next section we will show that the situation for the curricula
which receive the most penalty can be improved with max-min fair optimization
for 15 out of 21 instances.



\subsection {Max-Min Fair Optimization}
\label{sec:mmfopt}

In Section~\ref{sec:sa}, we presented algorithm \MMFSA for solving max-min
fair minimization problems. A crucial part of this algorithm is the
energy difference measure which determines how much worse a given solution is
compared to another
solution, i.e.\ the energy difference of the solutions. We evaluate the impact
of the three energy difference measures~\eqref{eq:elex},~\eqref{eq:ecw}
and~\eqref{eq:eps} on the performance of \MMFSA.

\begin{table}
	\caption{The performance of \MMFSA with $\Delta E = \Delta E_{\it cw}$ for
		different values of $\delta$. 		\label{tab:delta_comparison}}
$$
		\begin{tabular}{|l|c|c|c|c|} \hline
			$\delta$	&	$10^0$				&	$10^{-2}$	&	$10^{-3}$	&	$10^{-6}$\rule{0pt}{9pt}	\tabularnewline\hline
			$10^0$		&	--					&	\tt{02}		&	\tt{02,05}	&	\tt{02}		\tabularnewline\hline
			$10^{-2}$	&	\tt{10,19,20}		&	--			&	\tt{09}		&	\tt{19}		\tabularnewline\hline
			$10^{-3}$	&	\tt{01,10,19,20}	&	--			&	--			&	\tt{03,19}	\tabularnewline\hline
			$10^{-6}$	&	\tt{01,10,20}		&	--			&	--			&	--			\tabularnewline\hline
		\end{tabular}
$$
\end{table}

Our test setup was the following: For each energy difference function we
independently performed 50 runs with \MMFSA. The temperature levels were
determined experimentally, we set $\T_{max} =
5$ and $\T_{min} = 0.01$; the cooling rate $\alpha$ was set according
to~\eqref{eq:alpha}. In order to establish consistent experimental conditions
for fair optimization, we used a timeout, which was determined according to the
publicly available ITC2007 benchmark executable. On our machines (i7 CPUs running at 3.4GHz, 8GB
RAM), the timeout was set to 192 seconds. The \MMFSA algorithm was executed on
a single core. We generated feasible initial timetables for \MMFSA as a
preprocess using sequential heuristics proposed by \cite{Burke:07}. The soft
constraint violations were not considered at this stage. Since the preprocess
was performed only once per instance (not per run), it is not counted in the
timeout. However, the time it took was negligible compared to the timeout (less
than 1 second per instance).

Table~\ref{tab:delta_comparison} shows the impact of the parameter $\delta$ on
the performance of \MMFSA with energy difference measure $\Delta E_{\it cw}$.
For each pair of values we performed the one-sided Wilcoxon Rank-Sum test with
a significance level of $0.01$. The data indicates that for best performance,
$\delta$ should be small, but not too small. For $\delta=1$, \MMFSA beats the
other shown configurations on instance \texttt{comp02} but performs worse than
the other configurations on instances \texttt{comp10} and \texttt{comp20}. For
$\delta=10^{-6}$ the overall performance is better than for $\delta=1$, but
worse than for the other configurations. With $\delta=10^{-2}$ and
$\delta=10^{-3}$, \MMFSA shows the best relative performance. Thus, for our
further evaluation we set $\delta=10^{-3}$.

Table~\ref{tab:delta_e_comparison} shows the relative performance of
Algorithm~\MMFSA for the proposed energy difference
measures~\eqref{eq:elex},~\eqref{eq:ecw} and~\eqref{eq:eps}. The table shows for
any choice of two energy difference measures $i$ and $j$, for which instances
\MMFSA with measure $i$ performs significantly better than \MMFSA using measure
$j$. Again, we used the Wilcoxon Rank-Sum test with a significance level of 1
percent. The data shows that $\Delta E_{\it cw}$ is the best choice among the
three alternatives, since it is a better choice than $\Delta E_{\it lex}$
on all instances except \texttt{comp11} and a better choice than $\Delta E_{\it
ps}$ on five out of 21 instances. However, although $\Delta E_{\it cw}$ shows 
significantly better performance than the other energy difference measures, it
did not necessarily produce the best timetables on all instances. For the
instances \texttt{comp03}, \texttt{comp15}, \texttt{comp05} and \texttt{comp12}
for example, the best solution found with $\Delta E = \Delta E_{\it ps}$ was
better than with $\Delta E= \Delta E_{\it cw}$.

\begin{table}
\caption {The performance of \MMFSA
with energy difference measures $\Delta E_{\it lex}$, $\Delta E_{\it ps}$ and
$\Delta E_{\it cw}$.\label{tab:delta_e_comparison}}
$$
\begin{tabular}{|l|c|c|c|}
\hline
			$\Delta E$ &	$\Delta E_{\it lex}$	    	& $\Delta E_{\it ps}$	& $\Delta E_{\it cw}$  	\tabularnewline\hline
$\Delta E_{\it lex}$	& 	--			    	& --						& --						\tabularnewline\hline
$\Delta E_{\it ps}$	& all except \tt{01,06,08,11,17}		    & 	--		& \tt{18} 					\tabularnewline\hline
$\Delta E_{\it cw}$		& all except \tt{11} 	&  \tt{06,07,08,17,21}		& -- \tabularnewline\hline
\end{tabular}
$$
\end{table}

The data in  Table~\ref{tab:maxmin} shows a comparison of the sorted allocation
vectors of the known best solutions from the CB-CTT website by
\cite{ITC2007:web} with the best
solutions found
by the 50 runs of \MMFSA 
with $\Delta E = \Delta E_{\it cw}$.
First of all, for instances \texttt{comp01} and \texttt{comp11}, the allocation
vectors of the best existing solutions and the best solution found by \MMFSA
are identical. This means that \MMFSA finds reasonably good solutions despite
the certainly more complex fitness landscape due to max-min fair optimization.
We can also observe that the maximum penalty any curriculum receives is
significantly less for most instances and the penalty is more evenly
distributed across the curricula. This means that although max-min fair
timetables may have a higher total penalty, they might be more attractive from
the students' perspective, since in the first place each student notices an
unfortunate arrangement of his/her timetable, which is tied to the curriculum.
Furthermore, we can observe that if the total penalty of a known best solution
is rather low, then it is also good with respect to max-min fairness. For
several instances in this category, (\texttt{comp01}, \texttt{comp04},
\texttt{comp07}, \texttt{comp10} and \texttt{comp20}), the solution
found by \MMFSA is not as good as the known best solution with respect to
max-min fairness. We can conclude that if there is not much penalty to
distribute between the stakeholders, it is not necessary to enforce a fair
distribution of penalty.

\begin{table}
\caption{Comparison of the sorted allocation vectors of the known best solutions
from the website by \cite{ITC2007:web} with the allocation vectors found by
\MMFSA 
with respect to max-min fairness.\label{tab:maxmin}}
$$\small
\begin{tabular}{@{}|l|c|c|@{}}
\hline
Instance        & Known best solution                   & \MMFSA 
                                                          ($\Delta E = \Delta E_{\it cw}$)\tabularnewline
\hline
\tt comp01	& $5^{2},0^{12}$                                                & $5^{2},0^{12}$                            \tabularnewline
\tt comp02	& $\bf 4,2^{10},0^{59}$                                         & $4^{2},2^{31},1^{7},0^{30}$	\tabularnewline
\tt comp03	& $13,10^{3},9,7^{2},6^{4},5^{13},4,2^{6},0^{37}$               & $\bf 6^{4},4^{11},2^{22},1^{3},0^{28}$\tabularnewline
\tt comp04	& $7,6^{3},5^{4},4^{2},2,0^{46}$                                & $\bf 6^{4},4^{2},2^{4},1,0^{46}$\tabularnewline
\tt comp05	& $41^{2},36^{7},35^{5},32^{5},31^{6},30^{9},28,\ldots,2,0^{3}$ & $\bf 19^{2},18^{3},17^{3},16^{5},15^{2},14^{15},13^{5},\ldots,4^{8},3^{3},2$\tabularnewline
\tt comp06	& $12,7^{2},5^{4},2^{3},0^{60}$                                 & $\bf 12,4^{2},2^{30},1^{13},0^{24}$\tabularnewline
\tt comp07	& $\bf 6,0^{76}$                                                & $6,2^{23},1^{24},0^{29}$\tabularnewline
\tt comp08	& $7,6^{3},5^{4},4^{2},2^{2},0^{49}$                            & $\bf 6^{4},4^{2},2^{7},1^{5},0^{43}$\tabularnewline
\tt comp09	& $10^{5},9,7^{10},6^{6},5^{10},4,2,0^{41}$                     & $\bf 6^{9},4^{14},2^{17},0^{35}$\tabularnewline
\tt comp10	& $\bf 2^{2},0^{65}$                                            & $2^{19},1^{6},0^{42}$\tabularnewline
\tt comp11	& $0^{13}$                                                  & $0^{13}$\tabularnewline
\tt comp12	& $45,30^{14},28,27^{2},26^{5},25^{19},22^{4},\ldots,2^{2},0^{3}$& $\bf 10^{3},9^{6},8^{31},7^{7},6^{43},5^{2},4^{36},3^{2},2^{16},1,0^{3}$\tabularnewline
\tt comp13	& $8,7,6^{5},5^{7},4^{2},2^{3},0^{47}$                          & $\bf 6^{6},4^{4},2^{13},1^{6},0^{37}$\tabularnewline
\tt comp14	& $8^{4},7,5^{2},2^{6},0^{47}$                                  & $\bf 8^{4},4^{2},3,2^{18},0^{35}$\tabularnewline
\tt comp15	& $10^{3},9^{3},7,6^{4},5^{13},4,2^{7},0^{36}$                  & $\bf 6^{4},4^{11},2^{23},1^{2},0^{28}$\tabularnewline
\tt comp16	& $7^{2},5^{7},4,0^{61}$                                        & $\bf 4^{5},2^{16},1^{4},0^{46}$\tabularnewline
\tt comp17	& $10^{2},6^{3},5^{9},2^{4},0^{52}$                             & $\bf 10^{2},6^{2},4^{7},3,2^{25},1^{7},0^{26}$\tabularnewline
\tt comp18	& $17,15,14,13,11,10,9^{2},5^{19},2^{2},0^{23}$                 & $\bf 4^{20},2^{11},1^{5},0^{16}$\tabularnewline
\tt comp19	& $13,7,6^{4},5^{2},4,2^{7},0^{50}$                             & $\bf 6^{4},4^{6},2^{15},1^{14},0^{27}$\tabularnewline
\tt comp20	& $\bf 2^{2},0^{76}$                                            & $4^{5},3^{3},2^{31},1^{7},0^{32}$\tabularnewline
\tt comp21	& $12,11,10^{4},9,7^{4},6^{4},5^{12},4,2^{3},1^{2},0^{45}$      & $\bf 10,6^{4},5,4^{15},3,2^{36},1^{3},0^{17}$\tabularnewline
\hline                                                                          
\end{tabular}
$$
\end{table}

\subsection {The Tradeoff Between Fairness and Efficiency}
\label{sec:tradeoff}

We proposed the JFI-CB-CTT problem formulation in
Section~\ref{sec:fairtimetabling}, which allows us to investigate the tradeoff
between fairness and efficiency which arises in course timetabling. 
We can observe in column 4 of Table~\ref{tab:fairbest} that for all of the best solutions
from~\cite{ITC2007:web} the fairness index~\eqref{eq:jain} is greater than
$0.8$, i.\,e., the known best solutions are also fair for more
than 80 percent of the curricula. In order to solve the corresponding
JFI-CB-CTT instances, we use the multi-objective optimization algorithm AMOSA
proposed by \cite{BSMD:08} that is based on simulated annealing like
Algorithm~\MMFSA. Since we do not expect
from a general multi-objective optimization algorithm to produce solutions as
good as the best CB-CTT solvers, we will consider the following scenario to
explore the tradeoffs between fairness and efficiency: starting from the known
best solution we examine how much increase in total penalty we have to tolerate
in order to increase the fairness further. We will take as examples the six
instances with the highest total amount of penalty, \texttt{comp03},
\texttt{comp05}, \texttt{comp09}, \texttt{comp12}, \texttt{comp15}
and \texttt{comp21}. 

\begin{figure}
	\centering
	\begin{subfigure}{.4\textwidth}
		\includegraphics{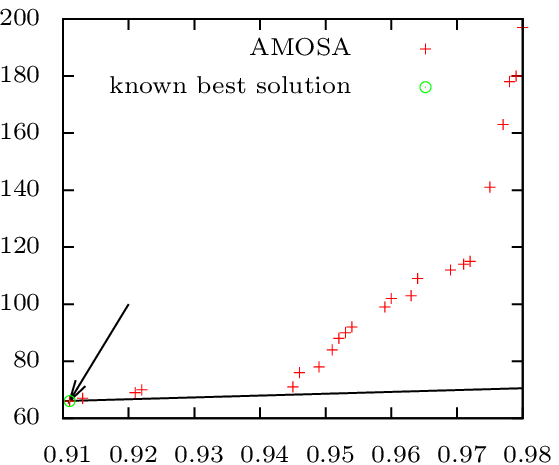}
	\end{subfigure}
	\hfill
	\begin{subfigure}{.4\textwidth}
		\includegraphics{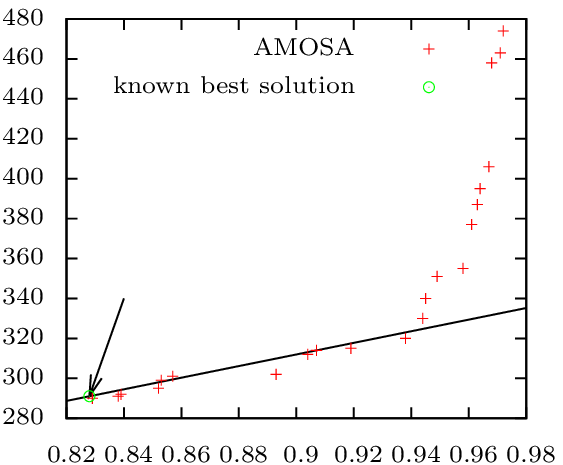}
	\end{subfigure}
	\begin{subfigure}{.4\textwidth}
		\includegraphics{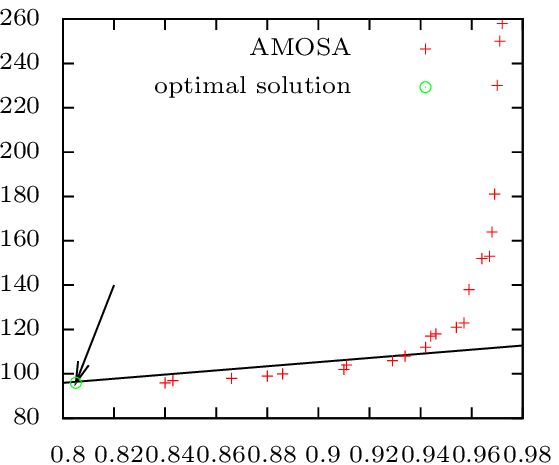}
	\end{subfigure} 
	\hfill
	\begin{subfigure}{.4\textwidth}
		\includegraphics{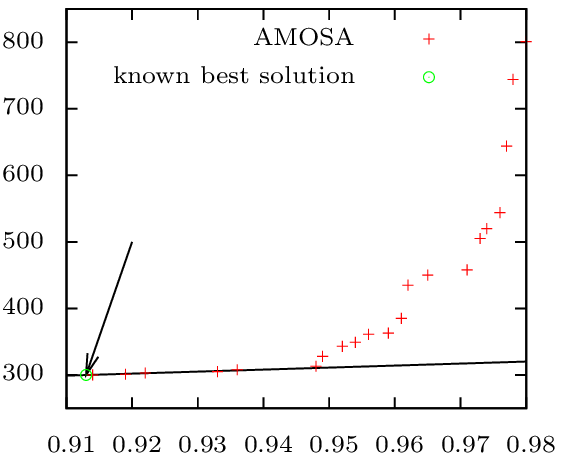}
	\end{subfigure}
	\begin{subfigure}{.4\textwidth}
		\includegraphics{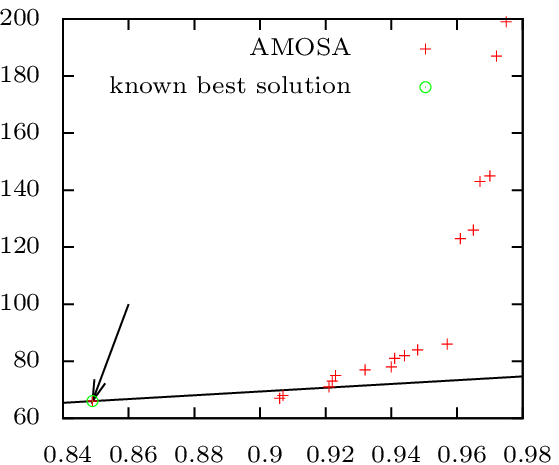}
	\end{subfigure}
	\hfill
	\begin{subfigure}{.4\textwidth}
		\includegraphics{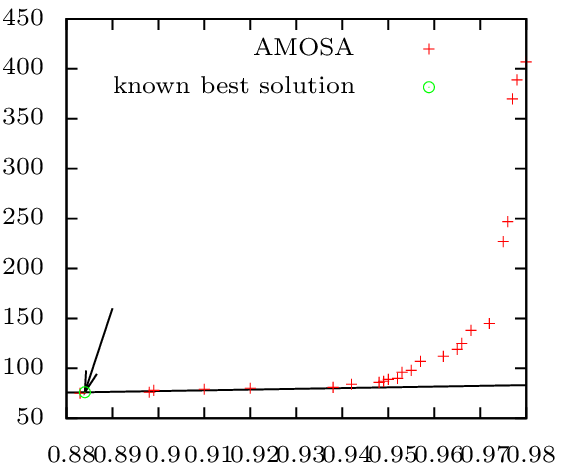}
	\end{subfigure}

	\caption{\label{fig:curves}Non-dominated solutions found by the AMOSA
	algorithm for the JFI-CB-CTT versions of instances
	\texttt{comp03}, \texttt{comp05}, \texttt{comp09}, \texttt{comp12}, \texttt{comp15} and
	\texttt{comp21}. All graphs show the fairness index on the horizontal axis and the amount of penalty on the vertical axis.\label{fig:amosa}}
\end{figure}

The temperature levels for the AMOSA algorithm were set to $\T_{max} = 20$ and
$\T_{min} = 0.01$; $\alpha$ was set according to~\eqref{eq:alpha} with a
timeout determined by the official ITC2007 benchmark. The plots in
Figure~\ref{fig:amosa} show the (Pareto-) non-dominated solutions found by
AMOSA. The arrows point to the starting point, i.e.\ the best available
solutions to the corresponding instances. For instances \texttt{comp05} and
\texttt{comp21} solutions with a lower total cost than the the previously known
best solutions were discovered by this approach. The plots show that the price
for increasing the fairness is generally not very high -- up to a certain level,
which depends on the instance. In fact, for \texttt{comp09} and
\texttt{comp21}, the fairness index can be increased by 3.5 percent and 1.4
percent, respectively, without increasing the total penalty at all. 

In Figure~\ref{fig:amosa}, the straight lines that go through the initial
solutions show a possible tradeoff between fairness and efficiency: the slopes
were determined such that a 1 percent increase in fairness yields a 1 percent
increase in penalty. For the instances shown in Figure~\ref{fig:amosa}, the
solutions remain close to the tradeoff lines up to a fairness of 94 to 97
percent, while a further increase in fairness demands a significant increase in total
cost. For the instances \texttt{comp05}, \texttt{comp09} and \texttt{comp15},
there are several solutions below the tradeoff lines. Picking any of the
solutions below these lines would result in an increased fairness without an
equally large increase in the amount of penalty. This means picking a fairer
solution might well be an attractive option in a real-world academic timetabling
context. For \texttt{comp05} for example, the fairness of the formerly best
known solution with a total penalty of 291 can be increased by 5.4 percent at
302 total penalty, which is a 3.8 percent increase.

In summary, improving the fairness of an efficient timetable as a
post-processing step seems like a viable approach for practical decision
making.  Using a very efficient solution as a starting point means that we can
benefit from the existing very good approaches to creating timetables with
minimal total cost and provide improved fairness depending on the actual,
instance-dependent tradeoff.

\section {Conclusion}

In this paper we introduced two new problem formulations for academic course
timetabling based on the CB-CTT problem model from track three of the ITC2007,
MMF-CB-CTT and JFI-CB-CTT. Both problem formulations are aimed at creating fair
course timetables in the setting of a university but include different
notions of fairness. Fairness in our setting means that the penalty assigned
to a timetable is distributed in a fair way among the different curricula. The
MMF-CB-CTT formulation aims at creating max-min fair course timetables while
JFI-CB-CTT is a bi-objective problem formulation based on Jain's fairness
index. The motivation for the JFI-CB-CTT formulation is to explore the tradeoff
between a fair penalty distribution and a low total penalty. 

Furthermore, we proposed an optimization algorithm based on simulated annealing
for solving MMF-CB-CTT problems. A critical part of the algorithm is concerned
with measuring the energy difference between two timetables, i.e., how much
worse a timetable is compared to another timetable with respect to max-min
fairness. We evaluated the performance of the proposed algorithm for three
different energy difference measures on the 21 CB-CTT benchmark
instances. Our results show clearly that the algorithm performs best with
$\Delta E_{\it cw}$ as energy difference measure.

Additionally, we investigated the fairness of the known best solutions of the 21
CB-CTT instances with respect to max-min fairness and Jain's fairness index.
These solutions were not created with fairness in mind, but our results show
that all of the solutions have a fairness index greater than $0.8$.  This means
they can be considered quite fair. Nevertheless, our results show that some improvements are possible
with respect to both max-min fairness and Jain's fairness index. The
timetables produced by our proposed \MMFSA algorithms are better than the
known best ones with respect to max-min fairness for 15 out of 21 instances.
Our investigation of the tradeoff between fairness and the total amount of
penalty using the JFI-CB-CTT problem formulation shows that the fairness of the
known best timetables can be increased further with only a small increase of
the total penalty.



\bibliographystyle{alpha}
\bibliography{paper}   

\end{document}